\documentclass[]{Liebert_Author}
\usepackage{amsmath,amssymb,amsfonts}
\usepackage{mathrsfs}
\usepackage{algorithmic}
\usepackage{graphicx}
\usepackage{textcomp}
\usepackage{subcaption}
\usepackage{siunitx}
\usepackage{multirow}
\usepackage{boldline}
\usepackage{footmisc}
\usepackage{tabto}
\usepackage{xcolor}
\usepackage{tablefootnote}
\usepackage{bm}

\newcommand{\boldmscr}[1]{\bm{\mathscr{#1}}}
\newif\ifshowcomments

\showcommentstrue

\ifshowcomments

\newcommand{\KSH}[1]{{\color{red}{#1}}} 
\newcommand{\KSR}[1]{{\color{green}{#1}}}
\newcommand\KSHsout{\bgroup\markoverwith{\textcolor{red}{\rule[0.5ex]{2pt}{0.4pt}}}\ULon}
\newcommand{\BJV}[1]{{\color{blue}{#1}}}
\newcommand{\requestedChange}[1]{{{#1}}}
\newcommand{\requestedChangeTWO}[1]{{{#1}}} 

\else
\newcommand{\KSH}[1]{}
\newcommand\KSHsout[1]{}%
\newcommand{\KSR}[1]{}
\newcommand{\BJV}[1]{}
\newcommand{\requestedChange}[1]{#1}
\newcommand{\requestedChangeTWO}[1]{#1}
\fi
\newcommand{\ttau}{\bm{\tau}}

\title{Model-based Optimization of  Anguilliform Swimming Gaits for Soft Robotic Applications} 

\author{
Brian Van Stratum$^{1}$,  James Gallentine$^{2}$,
Caleb Rucker$^{3}$,\\
Eric Barth$^{2}$,
 Jonathan E. Clark$^{1}$ and Kourosh Shoele$^{1\ast}$\\
     {$^1$ FAMU/FSU College of Engineering, Tallahassee, FL 32310}\\
    {$^{2}$ Mechanical Engineering, Vanderbilt University, Nashville Tennessee 37235}\\
    {$^{3}$Mechanical Engineering, The University of Tennessee, Knoxville, Tennessee 37996}\\
    {$^\ast$To whom correspondence should be addressed;}\\
          {E-mail:  kshoele@fsu.edu.}      
}

\begin{document} 
\maketitle 
\newpage
\begin{abstract}
In this paper, we introduce the Soft Lamprey-Inspired Dual Environment Robot (SLIDER) and a proper modeling and optimization procedure employed to design the robot. We represent the primary fluid environment actions—inertial effects, vortex forces, and viscous dissipation—using Lighthill's theory for large-amplitude elongated bodies. For structural design parameters such as internal pressure, tail size, and body stiffness, a fast, geometrically and materially nonlinear model is developed and validated. The fluid-structure interaction equations are solved implicitly with an efficient second-order box method. A pneumatic manifold robotic system is employed to actuate SLIDER in a quiescent water tank environment, allowing cross-comparison of computational and experimental results. We find that low-frequency swimming is dominated by resistant environmental forces, whereas higher-frequency swimming is primarily affected by inertial fluid forces.
\requestedChangeTWO{Using our efficient model alongside a genetic algorithm, we co-optimize a swimming control pattern and caudal fin design \requestedChange{(subject to SLIDER's climbing morphology) to achieve a tethered swimming speed of $21.7\pm0.4$ cm/s (0.59 Bl/s)}. Furthermore, we investigate the optimization procedure for a multimodal robot performing both swimming and climbing tasks.}
\end{abstract}
\newpage
\section{Introduction}
Designing soft, bio-inspired robots for diverse terrains is challenging because of the complex interactions between active actuation and the passive responses of flexible components, which require precise, adaptive computational models. Selecting a computationally feasible yet sufficiently accurate model is essential for optimizing autonomous robots, whether navigating icy moons or performing nanoscale drug delivery.

\requestedChangeTWO{Designing robots that interact with fluids presents a significant challenge, especially when multiple locomotion modes  such as swimming and climbing, must be integrated into a single platform. Multimodal systems require balancing conflicting physical demands, where structural features beneficial for one environment may degrade performance in another, requiring an analysis and design procedure that can accommodate different conditions. The complexity is especially pronounced in robots with swimming tasks and is further compounded by the fluid’s multiple timescales, which complicate modeling efforts. Bridging the gap between bio-inspired robots and their biological counterparts remains particularly difficult, given that animals inherently combine optimized morphology with adaptive control and cognitive functions \cite{beal2006passive}. Achieving comparable agility and robustness in robotic platforms that are employed for multiple environmental conditions necessitates efficient models that capture essential fluid–structure interactions at a low enough complexity for use in real-time design and control tasks \cite{bongard2006resilient}.}

\begin{figure}
    \centering
    \includegraphics[width = 0.5\textwidth]{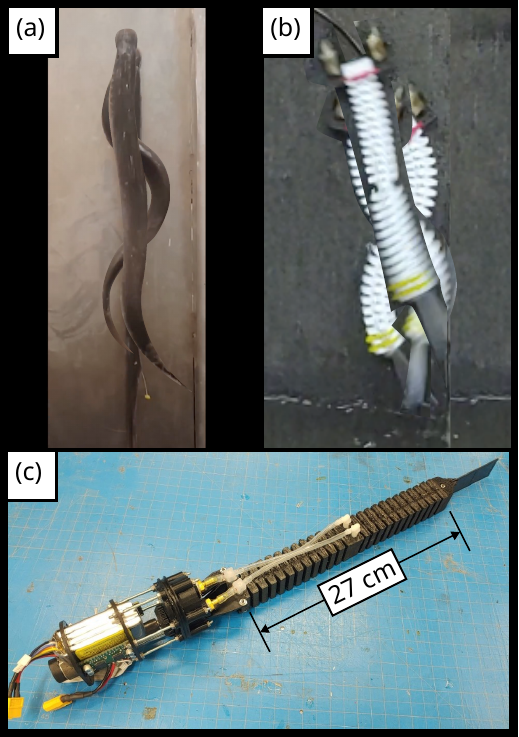}
    \caption{(a) Pacific lamprey climbing a vertical weir. Courtesy of Ralph Lampman (b) Composited video frames of soft robot SLIDER climbing carpeted wall with microspine array after exiting the water on power tether. (c) Prototype of the SLIDER robot's autonomous hardware.}
    \label{fig1:conceptImage}
\end{figure}

We seek an optimized design of the Soft Lamprey-Inspired Dual Environment Robot (SLIDER), depicted in Figure~\ref{fig1:conceptImage}. Our team's prior work describes SLIDER's development and climbing performance in detail \cite{gallentine2022multimodal}. The curse of dimensionality dictates that the combined design morphology and control space of any effective swimming-climbing robot, including SLIDER, is too large to optimize through experimental direct search alone. The need to balance performance across both locomotion modes—each governed by distinct environmental constraints—further complicates this optimization. Therefore, we propose a mid-fidelity mathematical model suitable for simulation-based design and control co-optimization under multimodal constraints.

Some designers have utilized low-complexity simplified kinematic models to study fluid interactions with slender swimming soft robots, using experimental kinematics as inputs to estimate fluid forces \cite{nguyen2020kinematic,nguyen2021anguilliform,nguyen2021evaluation}. While these models offer useful insights, their application in soft robot co-optimization is limited since kinematics are influenced by fluid forces and can only be determined experimentally. Alternatively, high-complexity computational fluid dynamics (CFD) models, which solve flow equations around the robot, provide greater accuracy but are computationally expensive and depend on predefined kinematic inputs, limiting their practicality for integrated design and control optimization \cite{feng2020body,cruz2020soft,shoele2016flutter}.

Recently, machine learning (ML) techniques have emerged as an alternative method for optimizing robotic body shapes, including swimmers, in simulated environments using resistive fluid models \cite{ma2021diffaqua,liu2022fishgym}. This group of approaches has also been enhanced by integrating more efficient ML-based fluid models \cite{nava2022fast}. Yet, ML techniques require extensive training, and their performance is reliable only in conditions closely matching those encountered during training, while first-principles models, similar to what is discussed here, can provide interpretability and robustness to conditions outside the training distribution. Unfortunately, most of the first-principles models are either expensive or simplified. Still, there is a need for validated dynamic fluid-structure interaction (FSI) models that capture key physical processes that influence the robotic platform's behavior.

Efficient dynamic models in particular enable the discovery of optimal adaptive locomotion strategies, offering insights into both body design and control strategies  \cite{bongard2006resilient}. Some moderate-complexity first-principle approaches have been proposed, notably drawing from M.J. Lighthill's foundational work on fluid-structure interaction (FSI), particularly his large amplitude elongated body theory (LAEBT) \cite{lighthill1960note,lighthill1971large}. LAEBT has been used to describe the swimming performance of segmented robots, with three-dimensional adaptations used to study the dynamics of slender, flexible objects in free-flow conditions \cite{porez2014improved, candelier2011three,leclercq2018vortex}. However, LAEBT has yet to be applied to reach optimized morphological design of soft-robotic swimmers and to perform controller codesign.

A key challenge in mid-complexity models is properly scaling flow forces, specifically the drag and added mass parameters, to match operating conditions. Hydrodynamic forces on a neutrally buoyant swimmer consist of resistive (drag) and inertial (added mass) components \cite{lighthill1960note}. Accurate parameter estimation, often requiring targeted experiments, is crucial for robotic applications \cite{sarpkaya2010wave}. Vortex-induced forces on oscillating bodies vary with amplitude and frequency, where either resistive or added mass forces dominate, as described by the Keulegan-Carpenter period and Reynolds numbers \cite{keulegan1956forces, sarpkaya1976vortex}.

 We aim to support control and design co-optimization by proposing a mid-complexity approach that combines LAEBT for fluid dynamics with a nonlinear beam model for structural physics to capture swimmer-fluid interactions. The chosen drag and added mass parameters are validated experimentally. This study introduces a forward dynamic FSI model and validates its results through propulsion tests conducted on SLIDER.
 
\requestedChangeTWO{
  This paper makes two primary new contributions. First, we present a novel adaptation of nonlinear large-amplitude beam theory, coupled with Lighthill’s Large-Amplitude Elongated Body Theory (LAEBT), to develop a fast, accurate dynamic model that captures the transfer function between pressure-controlled actuation and free-swimming motion in long, slender soft robots. Second, we validate this model by applying it to co-optimize the design and control of SLIDER and by considering the trade-off between the required optimality across different locomotion modes (here, climbing and swimming). }
  \requestedChangeTWO{Despite many aspects of the body having been optimized for climbing, SLIDER achieves strong swimming performance, highlighting the model’s effectiveness in handling trade-offs in multimodal robotic design. Using a model tailored to SLIDER’s multimodal design, we show that which morphological designs are good for both swimming and climbing. This demonstrates the effectiveness of the new modeling approach in managing the trade-offs required for multimodal robotic platforms.
 }

In Section \ref{Sec:01-model}, we present our computational model for a long, slender swimmer \requestedChangeTWO{as well as slender climber}. Section \ref{Sec:02-Methods} outlines the experimental, validation procedures, and co-optimization method. In Section \ref{Sec:03-Results}, we compare experimental and computational results based on the identified parameters. Finally, in Section \ref{Sec:04-conclusion}, we contextualize our findings within prior research and highlight improvements from using the Lighthill FSI model and a genetic algorithm.

\section{Computational Model}
\begin{figure}
	\centering
	\includegraphics[width=0.9\textwidth]{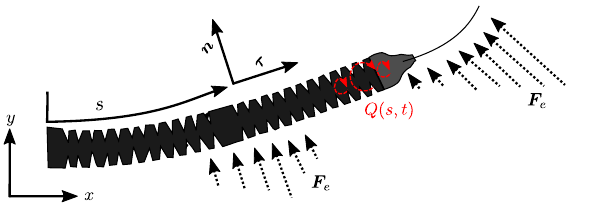}
	\caption{Free body diagram showing the body, adapter, and tail segments of SLIDER and its relation to the arc length parameter $s$  of the model. The local coordinate system $\mathbf{n}$ and $\mathbf{\ttau}$, internal actuation from pressure $Q(s,t)$, and the external forces $\bm{F}_e$. \requestedChange{Gravity acts into the page (i.e., perpendicular to the plane of motion).}} 
	\label{fig2:model}
\end{figure}

\label{Sec:01-model}
Computational models range from complex 3D fluid-structure simulations to simple kinematic-based analyses. We propose a mid-complexity model for the SLIDER robot, utilizing a nonlinear Euler–Bernoulli beam formulation based on the work of Tjavaras et al.\cite{tjavaras1996dynamics}. This model has been used to study the locomotion of fish and lampreys, including climbing using the jumping gate \cite{cheng1998continuous,zhu2011numerical,zhu2008propulsion}. Full details of the structural model and solver can be found in our previous work \cite{vanstratum2023Effects}. Here, we focus on presenting the model's key components. This includes the use of an Euler-Lagrangian dual coordinate system for formulating motion, encompassing the global coordinate system ($x$, $y$, $z$) and a local reference system defined by unit vectors $\ttau$, \textbf{n}, and \textbf{b} representing tangential, normal, and bi-normal (out of plane) directions, respectively.
   
\requestedChangeTWO{Dynamic equations and the centerline compatibility condition are derived as:
\begin{align}
m\left( {\frac{{\partial {\bm{V}}}}{{\partial
t}} + {\bm{\omega }} \times {\bm{V}}} \right) &= \frac{{\partial
{\bm{T}}}}{{\partial s}} + {\bm{\Omega }} \times {\bm{T}} +
\left( {1 + \varepsilon } \right) {{{\bm{F}}_e}
}  \\  
\frac{{\partial {\bm{M}}}}{{\partial s}} +
{\bm{\Omega }} \times {\bm{M}} &=\,-\left( {1 + {\varepsilon}}
\right)^3\ttau  \times {\bm{T}} -\left( {1 + {\varepsilon}}
\right)^2, \\
\frac{{\partial \varepsilon }}{{\partial
t}}{{\ttau }} + \left( {1 + \varepsilon } \right){\bm{\omega
}} \times {{\ttau }} &= \frac{{\partial {\bm{V}}}}{{\partial
s}} + {\bm{\Omega }} \times {\bm{V}}, 
\label{a1-eq5}     
\end{align}
Here, $m$ is the mass per unit length and $s$ is the arc length from the tail. The vectors ${\bm{V}}(s,t)$, ${\bm{\omega }}(s,t)$, ${\bm{T}}(s,t)$, and ${\bm{M}}(s,t)$ represent the translational velocity, angular velocity, internal forces, and internal moment, respectively, each resolved in the local $\{\ttau, {\bm{n}}, {\bm{b}}\}$ frame. The axial strain is $\varepsilon (s,t)$, while ${\bm{\Omega }}(s,t)$ is the Darboux vector measuring material torsion and curvatures. External force density ${\bm{F}}_e$ (a nonlinear function of ${\bm{V}}$ and ${\bm{\omega}}$) and internal moment density ${\bm{Q}}$ from pneumatic actuation will be defined in subsequent sections.

Internal forces and moments are related to strain and curvature via linear constitutive relations $T_\tau = EA\varepsilon$, $M_\tau = GJ\Omega_\tau$, and $M_{n,b} = EI\Omega_{n,b}$, where $EA$, $GJ$, and $EI$ represent axial, torsional, and bending stiffness. To account for material strain storage and pneumatic dampingin SLIDER robot, we employ a linear viscoelastic framework with a complex modulus $E = E_{el}(1 + \alpha \partial/\partial t)$, where $\alpha$ is the effective damping ratio. We assume $EI$ is invariant and deformations are small enough to prevent self-collision. The final system of equations is expressed compactly.

\begin{equation}
\frac{{\partial {\bm{Y}}}}{{\partial s}} +
{\boldmscr{H}}\frac{{\partial {\bm{Y}}}}{{\partial t}} + {\boldmscr{P}}
= 0, \label{a1-eq7}    
\end{equation} 
where ${\bm{Y}} = {\left[ {\varepsilon
\;{T_n}\;{T_b}\;{V_\tau }\;{V_n}\;{V_b}\;{\beta _0}\;{\beta _1}\;{\beta
_2}\;{\beta _3}\;{\Omega _\tau }\;{\Omega _n}\;{\Omega _b}}
\right]^T}$ is the space-time continuous space state variables. Here $\beta_0,\beta_1,\beta_2,\beta_3$ are the Euler parameters (unit quaternion) describing the orientation of the local material frame $\{\ttau, {\bm{n}}, {\bm{b}}\}$; the angular velocity $\bm{\omega}$ is related to their time derivatives through the standard quaternion kinematic relation. The resulting system is expressed in compact form, as described in \cite{tjavaras1996dynamics}, with the addition of environmental forces discussed later.

Thirteen boundary conditions are imposed. For free swimming, the free tail maintains $\beta_0 = 0, \bm{T} = 0,$ and $\bm{M} = 0$; for head-fixed cases, the head requires zero deflection and slope. For climbing or head-fixed cases with a rigid link of length $l_p$ attached perpendicular to the head, one end of the link is constrained as a non-holonomic pin while the other end remains free written in body attached  coordinate at the head as, $
\mathbf{r}_\text{pin} = \mathbf{r}_\text{head} + [0,\pm l_p/2,0]^T
$. The non-holonomic constraints at the pinned can be written as:
\begin{align}
\bm{V}_\text{pin} &= \bm{V}(0,t) + \bm{\Omega}(0,t) \times (\mathbf{r}_\text{pin} - \mathbf{r}_\text{head}) = 0. \\
{M}_{b,\text{pin}} &= EI\,\Omega_b(0,t) + \bm{T}(0,t) \times (\mathbf{r}_\text{pin} - \mathbf{r}_\text{head})\cdot \bm{e}_b = 0.
\label{BC2}
\end{align}

To solve Eq. (\ref{a1-eq7}), we distribute points \(s_k\) \((k = 1, \cdots, N_p)\) uniformly along the unstretched length, dividing the body into \(N_p-1 \) segments of length \(\Delta s\). An implicit box method advances the equations in time from \(t_{i-1}\) to \(t_i = t_{i-1} + \Delta t\) \cite{tjavaras1996dynamics,wendroff1959centered}. This method is second-order accurate in space and first-order in time, employing a lower-order time integration scheme for numerical stability. \requestedChange{Our method's computational complexity is \(O(N_p)\) per time step, and a grid convergence study shows \(N_p = 51\) is adequate for this analysis. In order to practically quantify compute time, we ran 60,000 time steps to achieve 10 seconds of simulated swimming time on a typical single CPU core of an \SI{2.7}{\giga\Hz} Intel(R) Xeon(R) Platinum 8168. This simulation required 76 seconds, achieving a compute time of \SI{~12.0}{\milli\second} per time step. }
}

\subsection{Fluid Forces}
We adopt the fluid force models proposed by Lighthill \cite{lighthill1971large} and Candelier et al. \cite{candelier2011three}, where the total fluid force ${\bm{F}}_e$ is partitioned into an added mass term ${\bm{F}}_a$ and a viscous drag term ${\bm{F}}_v$. This decomposition allows for a distinct treatment of inductive and resistive effects acting on the body:

\begin{equation}
\label{eqn reactive terms}
    \bm{F}_a = -m_a\frac{\pi\,\rho\,H^2}{4}
 \left(
     \frac{\partial (u_n \, \mathbf{n})}{\partial t}
    -\frac{\partial}{\partial s}(u_{\tau}\,u_n\,\mathbf{n})
    +\frac{1}{2}\,\frac{\partial\,( u_n^2 \mathbf{\ttau})}{\partial s}
 \right)   
 \end{equation}
Where $H$ is the body’s thickness in the binormal direction fixed along the $z$-axis  (Figure \ref{fig2:model}), $\rho$ is water density, and $m_a$ is the added mass coefficient. The relative velocity $\mathbf{u}=[u_{\tau},u_n,u_b]^T$ within the body’s local reference frame is defined as $\mathbf{u} = \bm{V} - U_{\text{swim}} \bm{i}$, where $U_{\text{swim}}$ is the steady swimming velocity in the negative $x$-direction.  The inductive force model in Eq. \ref{eqn reactive terms} captures both temporal fluid acceleration and acceleration induced by body reorientation. At high Reynolds numbers, the resistive force $\bm{F}_v$ is dominated by frictional and pressure-induced resistance from vortex separation. Per Eloy et al. \cite{eloy2012origin} and Taylor \cite{taylor1952analysis}, this drag is assumed to act normal to the centerline and is defined as:
 
\begin{equation}
\label{eqn resistive term}
	\bm{F}_v = -\frac{1}{2}\rho\,C_d\,H|u_n|\,u_n\,\mathbf{n}
\end{equation}
where $C_d$ is the dimensionless drag coefficient. The force model is solved implicitly alongside the body momentum equations (Eq. \ref{a1-eq7}). 
\subsection{Actuation}
Experimental characterization revealed that the robot's performance is governed by a first-order hydraulic pressure response, necessitating a mathematical description of these dynamics. Initial modeling attempts using idealized sawtooth and square-wave pressure waveforms produced internal actuation torques that were not observed experimentally. These idealized inputs introduced higher-order oscillatory modes and parasitic behaviors into the model, which, in the physical system, are naturally attenuated by hydraulic delays. Consequently, accounting for the pneumatic pressure response is critical for model fidelity. Furthermore, modeling this first-order system allows for a quantitative assessment of the hardware’s operational bandwidth. We represent the actuation pressure dynamics as:

\begin{equation}
    p_\text{eff}(t) = p_\text{set}(t) -e^{-t/k}\left(p_\text{set}(t) -p_0\right)
\end{equation}
Here, $p_\text{eff}$ is the effective pressure that results in net imposed internal torque of  $\bm{M}_I=K_Q\,p_\text{eff}\,\bm{b}$ at the posterior adapter with $K_Q$ being the distributed torque parameter. This factor was identified through calibration and parametric identification as explained in Section \ref{A}. $k$  is the manifold time constant for step pressure rise. $p_\text{set}(t)=\hat{p}_\text{set} H(t)$ is the time-dependent gauge pressure applied to the manifold with amplitude of $\hat{p}_\text{set}$ and a periodic squarewave actuation waveform of $H(t)$.
 The initial pressure $p_0$ is computed as a function of the driving frequency $f$ as, 
 $
    p_0 = p_\text{set}\left(1-e^{-f/(2\,k)}\right).
$
In this way, we account for the actuation bandwidth relevant to a hardware instantiation. Finally, the localized actuation torque, $\bm{M}_I$  at position $q$ along the robot is represented in the continuous beam equation as a distributed moment on the centerline of the model using a regularized Delta function ($\delta$) proposed by Peskin \cite{peskin2002immersed}.
\[
\bm{Q}(s,t) = \int_L {\bm{M}_I}(q,t)\,\delta(s-q)\,dq
\]

\section{Experimental Methods}
\label{Sec:02-Methods}
To validate the LAEBT beam model, we compared the experimental and computational reaction forces at the foremost point of the body with forces measured in controlled head-fixed experiments on the robot. We also compared the predicted and measured body shape kinematics.

\subsection{Parameter Identification}
\label{A}
SLIDER consists of four independent soft actuators arranged in two antagonistic pairs (Fig. \ref{fig3:exp setup} (b)). A detailed description of its construction and climbing performance is available in prior work \cite{gallentine2022multimodal, mosadegh2014pneumatic}. Mass was measured with a gram scale. 
\requestedChange{$EI$ was determined by applying a \SI{490}{\milli\newton} lateral load to the tip and measuring deflection $\Delta y$.}
The relationship $EI = \tfrac{P\,d^3}{3\,\Delta y}$ was tested for various loads, confirming the adequacy of a linear model. The damping parameter $\alpha$ was calculated by releasing a known load and recording the tip displacement history with a Vicon system at \SI{300}{\Hz}, using a Newton search to determine the decay ratio. A \SI{70}{\mm} tail at the trailing end of SLIDER had its $EI$ measured as \SI{5800}{\newton\mm^2}.

\begin{table}[h]
    \centering
    \caption{Parameters for Head Fixed Experimental and LAEBT-Beam Model}
    \begin{tabular}{|l|c|c|c|}
    \hline  
    Variable                  & Symbol    & Value                       &  Unit                         \\ \hline
    Linear Mass Density  &  $m$      &    0.7                      & \SI{}{\kg\per\meter}          \\
    Body Stiffness            &  $EI_b$     &  38000                    & \SI{}{\newton\milli\meter^2}        \\
    Tail Stiffness            &  $EI_t$     &  5800                     & \SI{}{\newton\milli\meter^2}        \\
    Damping Ratio             &  $\alpha$ &  5  $\times 10^{-6}$        & \SI{}{\second}                \\
    Submerged Height          &  $H$      & 30       & \SI{}{\mm}                       \\
    Pressure Torque Ratio            &  $K_Q$ & 479         & \SI{}{\mm^3}                          \\
    Torque from Set Press.     &  $Q_\text{set}$ & 175                 & \SI{}{\milli\newton}                  \\
    Manifold Time Constant    & $k$       & 87                     &\SI{}{\milli\second}                 \\
     \hline
    Water Density        & $\rho$    &  1000       & \SI{}{\kg\per\meter^3} \\ 
    Drag Coefficient     & $C_{d}$   &  3.0-4.0    &                       \\
    Added Mass           & $m_{a}$   &   0.0-1.0   &                       \\
    Low Break Frequency  & $f_\text{low}$ &  2          &\SI{}{\Hz}             \\
    High Break Frequency & $f_\text{high}$&  3          &\SI{}{\Hz}             \\
     \hline
    \end{tabular}
    \label{Tab:SimDATA}
 \end{table}

Physical pressure and computational distributed torque are related by the ratio $K_Q$. This parameter was determined by measuring static tip displacement of \SI{102}{\mm} while applying \SI{45}{\kPa} pressure in both chambers 1 and 2. 
Initial computational modeling showed that tail displacement amplitude was inversely related to the drag parameter $C_d$ and minimally affected by $m_a$, while thrust increased monotonically with $m_a$. Thus, a unique combination of $m_a$ and $C_d$ can be identified to match both the measured mean thrust and tail lateral excursion amplitude. To account for resistively and reactively dominated regimes, as demonstrated by Kuelegan \cite{keulegan1956forces} and Sarpkaya \cite{sarpkaya2010wave}, we propose a model parameterized by the actuating frequency by using this interpolation function between low- and high-frequency limits,
\begin{equation}
\mathcal{I}(f;a_\text{low},a_\text{high})=
\begin{cases}
a_\text{low}, & f\le f_\text{low},\\
a_\text{low}+\dfrac{f-f_\text{low}}{f_\text{high}-f_\text{low}}\left(a_\text{high}-a_\text{low}\right),
& f_\text{low}\le f\le f_\text{high},\\
a_\text{high}, & f\ge f_\text{high},
\end{cases}
\end{equation}
to define added mass and drag coefficients as
\begin{equation}
m_a(f)=\mathcal{I}(f,m_{a\,\text{low}},m_{a\,\text{high}}), \qquad
C_d(f)=\mathcal{I}(f,C_{d\,\text{low}},C_{d\,\text{high}}).
\end{equation}
Here, various parameters associated with the robot's hydrodynamic forces are listed in Table \ref{Tab:SimDATA}.

\subsection{Tank Tests with the Fixed Head}
\label{Methods-HeadFixed}
\begin{figure}
    \centering
	\includegraphics[width=1.0\textwidth]{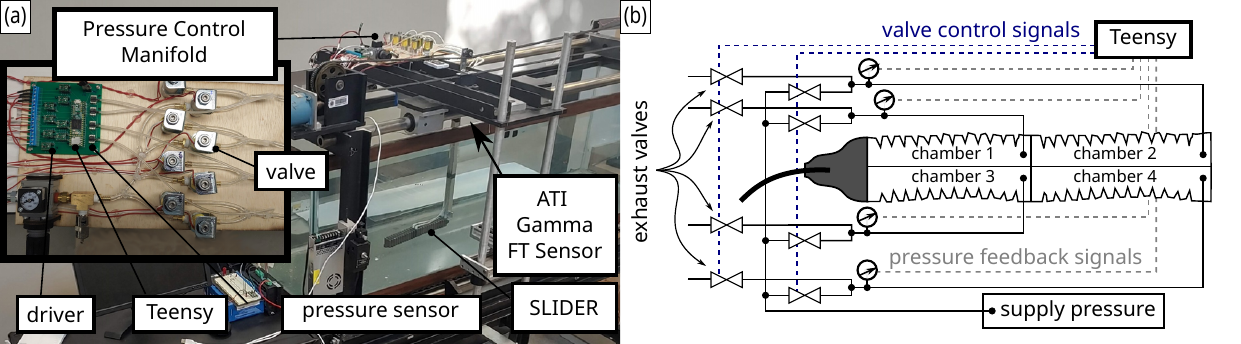}
	\caption{(a)Experimental setup including pressure control manifold on air bearing gantry, ATI force torque sensor, and fixturing for SLIDER Robot.\requestedChange{ (b) Pressure control diagram. Two desired relative pressure profiles are controlled via PWM control of the driver gates. Chambers 1 and 3 form an antagonistic pair, likewise 2 and 4.}
	} 
	\label{fig3:exp setup}
\end{figure}

Figure \ref{fig3:exp setup} inset depicts the pressure control manifold for actuating SLIDER's four chambers. Compressed air at \SI{760}{\kPa} is regulated and delivered to two iQ proportional solenoid valves per channel. Pressure is sampled using a Honeywell ABPDRNN100PGAA5 sensor and controlled via a PJRC Teensy, which adjusts valves using a DRV8838 motor driver (part no. 2990). The control code is available online \cite{VanStratum2023Pressure}.

For this study a \SI{185}{\centi\meter} $\times$ \SI{46}{\centi\meter} $\times$ \SI{58}{\centi\meter} tank was filled with water to \SI{25}{\cm}. An ATI Delta six-axis force-torque sensor (FT) was mounted to the bottom of a moving platform on metal rails above the water. SLIDER was connected to the FT with a one-inch aluminum channel. Figure \ref{fig3:exp setup} (a) shows the main components of the tank experiment setup.

Thrust was measured from 0.4Hz to 7.0Hz in 0.2Hz steps, with each test run for 20 cycles. Force data were sampled with an ATI Delta at 10 kHz, and video recorded at 60 FPS (1920$\times$1080), with select tests at 400 FPS (768$\times$288). Tests were automated via serial communication with the Teensy, and MATLAB scripts identified steady thrust production after 3 tailbeat cycles. Mean force and standard deviation were calculated for each stride based on the driving frequency. The Matlab code used is available online \cite{VanStratum2023Pressure}.

\subsection{Tank Tests - Tethered Swimming Tests}
Experimental assessments of SLIDER's locomotion were conducted by removing the head constraint and adding stabilizing outriggers to prevent roll. A vision-based tracking system utilizing OpenCV \cite{opencv_library} monitored blue markers to determine swimming speed in real time. To maintain a tethered but unbiased environment, the gantry executed a trapezoidal velocity profile; the specific speed was determined via a Newton search to ensure the robot remained stationary relative to the camera (zero pixel displacement) during steady-state swimming. This setup enabled two primary sweep studies: one investigating the impact of passive tail length and another examining the phase delay between front and rear chamber actuation.

We observed that passive tail length and shape differ among slender swimmers in nature, motivating a systematic assessment of SLIDER’s model by testing tails of equal cross-section but varying lengths in \SI{4}{\cm} increments. Since many swimmers propagate waves from head to tail, we explored the phase delay between front- and rear-chamber actuation by varying phase shifts during free-swimming tests.

A genetic algorithm was employed to optimize SLIDER's locomotion efficiency. The design space included tail length $L$, stiffness $EI$, and control parameters: actuation amplitudes for front and rear chambers ($P_\text{head}$, $P_\text{tail}$), phase lag $\phi$, and actuation frequency $f$. Following Katoch et al. \cite{katoch2021review}, a genetic algorithm with fitness as a linear function of output was used, with mutations decreasing over 100 generations. \requestedChangeTWO{A similar approach is also used to optimize a multimodal robot based on efficiecy for both climbing and swimming tasks, as discussed later.}

\section{Results}
\label{Sec:03-Results}
\subsection{Head Fixed Model Validation}
\begin{figure}
    \centering
    \includegraphics[width = 0.5\textwidth]{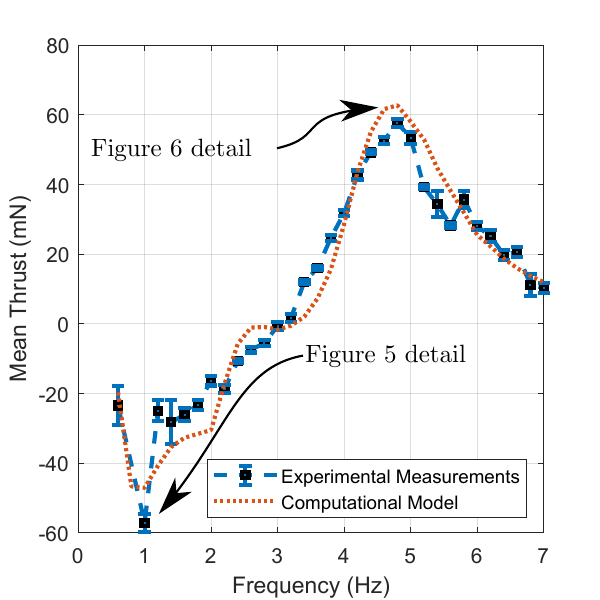}
    \caption{Experimental vs computational thrust data as a function of frequency for SLIDER robot. Error bars on the experimental data represent one standard deviation ($n=17$) tail beats. Tail height for this case is \SI{30}{mm} and the stiffness $EI$ \SI{5800}{\newton\mm^2}. Computational modeling shows the steady mean clamped thrust production at each frequency.}
    \label{fig4:Results-FreqSweep}
\end{figure}

To explore the interaction of fluid and elastic properties with gait kinematics and thrust repeatability, we analyze forces generated by the robot's periodic tailbeat. Figure \ref{fig4:Results-FreqSweep} illustrates the mean thrust produced as a function of tailbeat frequency, with error bars indicating one standard deviation. Thrust reaches a minimum at approximately 1 Hz, increases with frequency, and begins to roll off at about \SI{20}{\milli\newton\per\Hz} after peaking near \SI{60}{\milli\newton} at \SI{4.8}{\Hz}. \requestedChange{Comparing residuals of the computational model's thrust force to the experimental values gives a $R^2$ value of 0.95.}

\begin{figure}
    \centering
    \includegraphics[width = 1.0\textwidth]{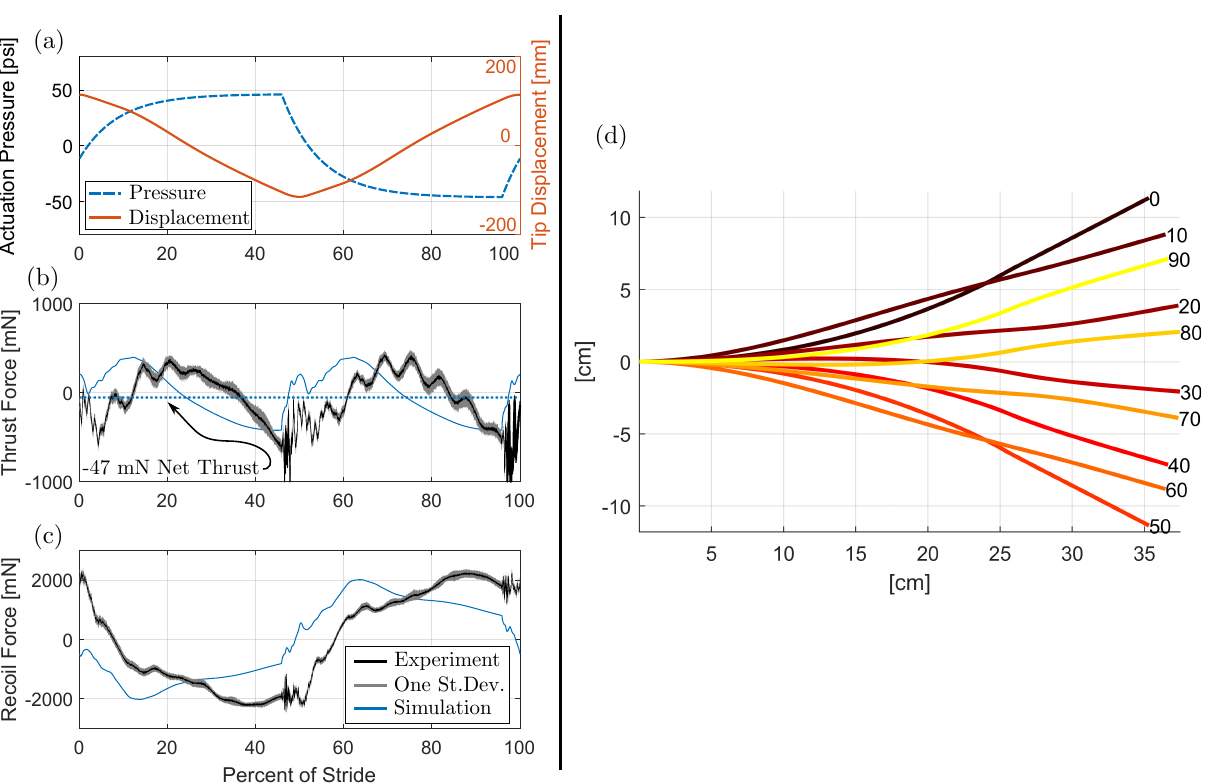}
    \caption{Computed and experimental time series data for the \SI{1}{\Hz} case ($M_a = 0.0$ and $C_d = 4.0$). (a) Actuation pressure and lateral excursion displacement of the tail tip. (b) Measured and computational reaction forces in the $x$ or thrust direction. (c) Measured and computational reaction forces in the $y$ or recoil direction. The gray-shaded regions of (b) and (c) show one standard deviation of the experimental result at that point in the stride. (d) Computational body trajectory of SLIDER actuated at one Hz. Numbers 0 - 90 at the tip of the tail indicate the percent of stride. Note the quick change in body concavity between 0 and 10 and 50 to 60, and the effect on tail angle. See the supplemental video for a video comparison of the experimental shapes.}
    \label{fig5:one Hz dynamics}
\end{figure}

Figure \ref{fig5:one Hz dynamics} presents the 1 Hz case dynamics, showing tail displacement and driving pressure (a) with corresponding thrust force (b). The reaction force (solid blue) is compared against the mean experimental thrust (black) and its standard deviation (grey). Experimental recoil forces exceed 1000 mN for 60\% of the cycle. As shown in Figure \ref{fig5:one Hz dynamics}d, substantial tail-tip angular changes contribute to high Reynolds number flow and increased drag via vortex formation.

\begin{figure}
    \centering
    \includegraphics[width = 1.0\textwidth]{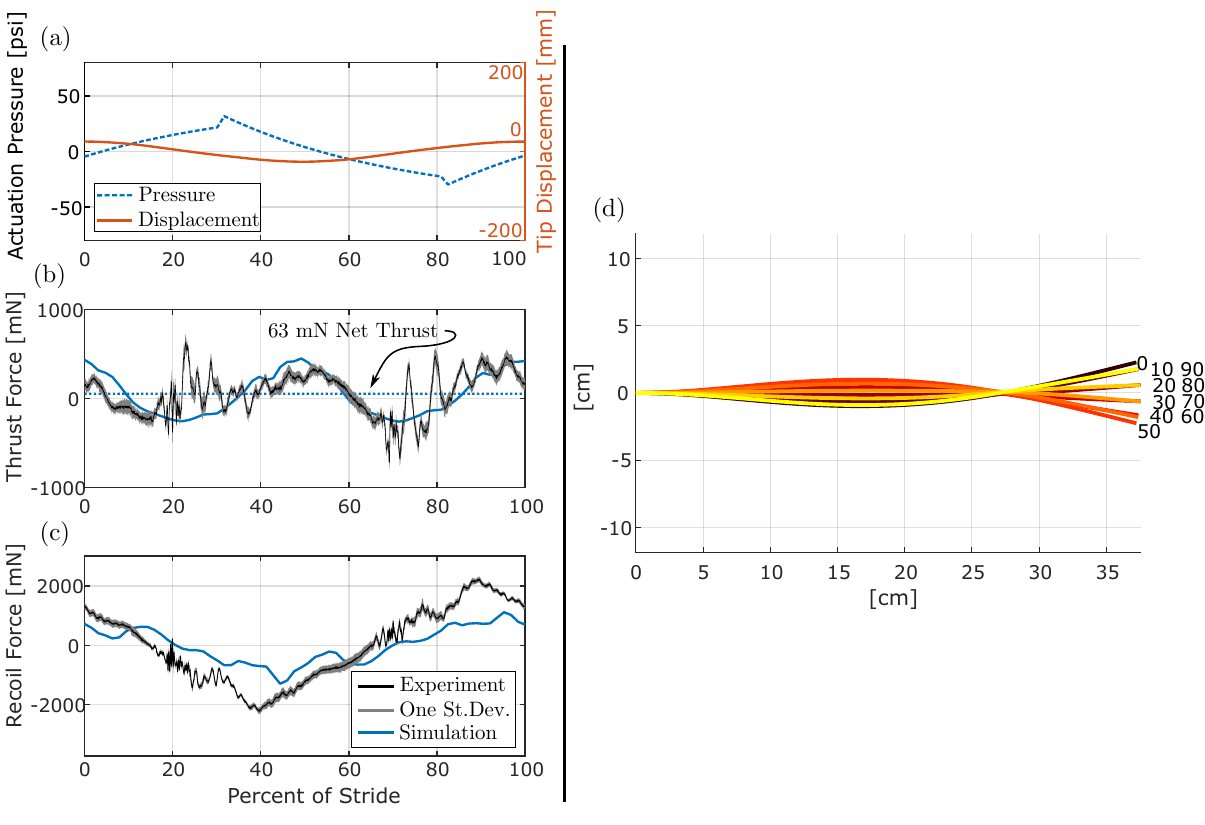}
    \caption{Computed and experimental time series data for the \SI{4.8}{\Hz} case. ($M_a = 1.0$ and $C_d = 3.0$). (a) Actuation pressure and lateral excursion displacement of the tail tip. Note the pressure reduction compared to Fig. \ref{fig5:one Hz dynamics} (a). (b) Measured and computational reaction forces in the $x$ or thrust direction. (c) Measured and computational reaction forces in the $y$ or recoil direction. The gray-shaded regions of (b) and (c) show one standard deviation of the experimental result at that point in the stride. (d) Computational body trajectory of SLIDER actuated at 4.8 Hz. This mode shape is characteristic of the frequencies that produce positive mean thrust. See the supplemental video for an overlay comparison of experimental body shapes.}
    \label{fig6:4p8 Hz Body Dynamics}
\end{figure}

Figure 6 captures the 4.8 Hz case, with Subfigure (a) showing reduced tail displacement due to pneumatic limitations. Subfigure b compares computational and experimental thrust data, revealing a more sinusoidal thrust profile at this frequency, leading to positive net thrust. In Subfigure (c), the recoil force shape becomes ramp-like, also exceeding 1000 mN for about 40\% of the cycle. Subfigure (d) indicates that the body deflection primarily follows its second mode shape, enhancing inertial effects while reducing flow separation. This transition to inertial-dominant flow is crucial for generating net thrust, enabling effective swimming.

\subsection{Free Swimming Studies}
\begin{figure}
  \centering
    \includegraphics[width= 1.0\textwidth]{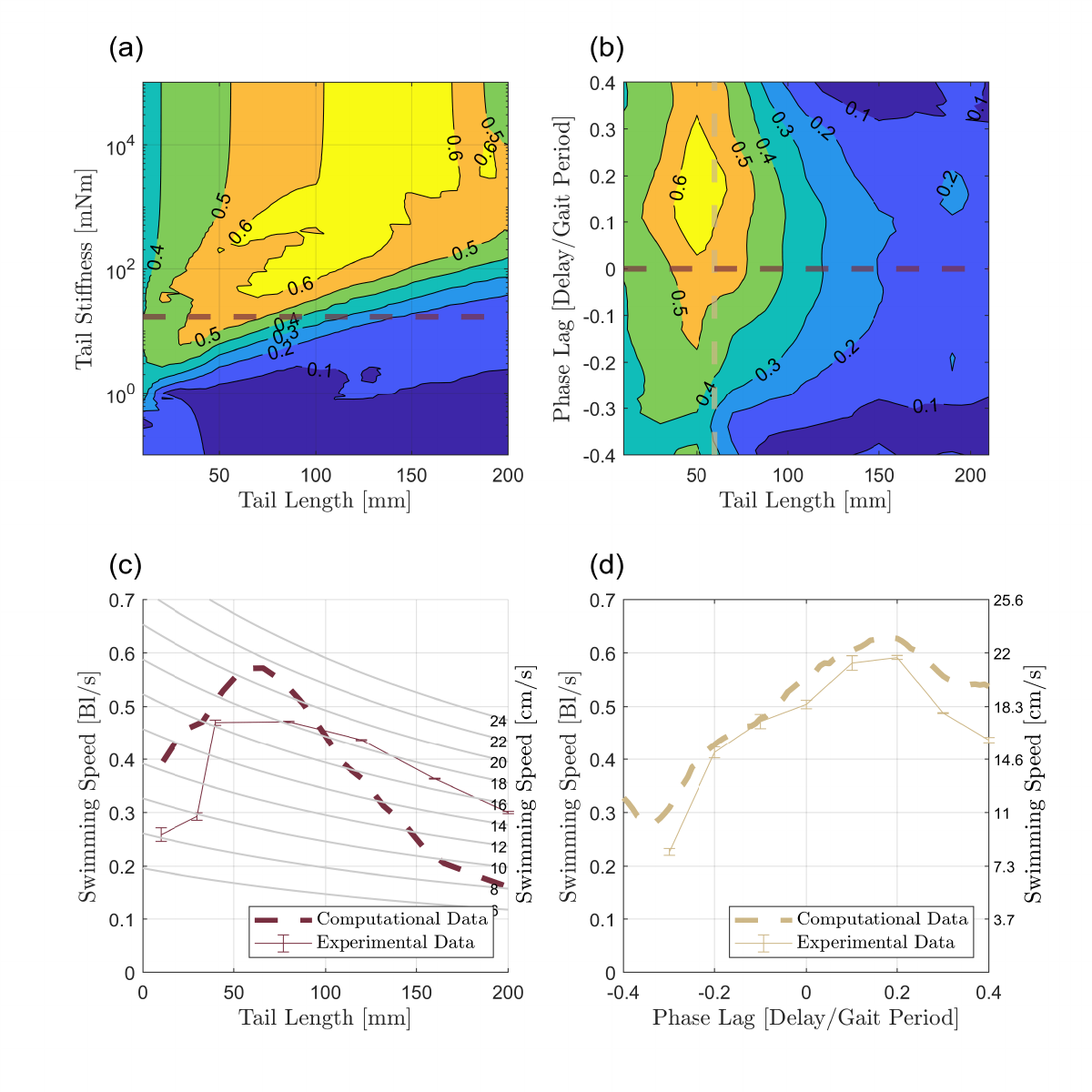}
    \caption{(a) Parameter variational study for tail length and stiffness. Contours depict free swimming speed in Body lengths per second. The garnet dashed line represents the slice of the parameter space selected for an experimental study. (b) Parameter variational study for tail length and phase. The Gold dashed line represents the slice of the parameter space selected for an experimental study. (c) Comparison of the computational and experimental swimming speeds as a function of tail length for a constant stiffness (EI = 17 mN$\cdot$m$^2$). (d) Comparison of the computational and experimental swimming speeds as a function of phase shift for a constant length (L = 60mm). The peak measured speed here is $21.3\pm0.1$}
    \label{fig:7 Results-LengthVStiffness}
\end{figure}

Figure \ref{fig:7 Results-LengthVStiffness} (a) shows the results of a computational sweep of tail stiffness versus length. This study was conducted at 5 Hz, which is very close to the frequency that maximizes thrust production for the head-fixed case. The LABET-Beam model predicts that for every tail length $L$, there is a stiffness $EI$ that produces peak thrust. Beyond this point, increases in stiffness result in a slight reduction in swimming speed. Figure \ref{fig:7 Results-LengthVStiffness} (c) illustrates swimming speeds as a function of tail length for a constant stiffness $EI = 17$ mN$\cdot$m$^2$. Although the model overpredicts the swimming speed, there is an acceptable similarity between the computational and experimental trends. Additionally, the computational and experimental results agree on the tail length that produces peak swimming speed.

The phase delay between front- and rear-chamber actuation is related to the induction of a traveling curvature wave from head to tail in a long, slender swimmer, a phenomenon known to be effective for swimming. Figure \ref{fig:7 Results-LengthVStiffness} (b) shows how the tethered swimming velocity of the SLIDER robot changes as a function of the phase lag between the head and tail actuation chambers and the tail length. Here, all other parameters are fixed at their representative values provided in table \ref{Tab:SimDATA}. The computational model predicts that for relatively short tails, phase delay increases swimming speed up to approximately $15\%$ of the gait period. Conversely, if the tail chamber leads the head actuation, as in the case of negative delay numbers, the swimming speed is reduced. We compare the computational and experimental results for a particular choice of tail length highlighted in Figure \ref{fig:7 Results-LengthVStiffness} (d), where an excellent agreement is found between predicted swimming speed and experimental observations. The computational model predicts that longer tail lengths do not benefit from the effect of delay to the same degree as shorter tails.

\subsection{Genetic Algorithm Co-Optimized Swimmer}
The Genetic Algorithm converged to an optimized swimmer design that predicted a swimming speed of 23.2 cm/s. The algorithm converged on this optimal solution after 30 generations or 3,360 individual computational runs. The optimal configuration from the GA is 
\begin{table}[h]
\centering
\begin{tabular}{c c c c c c}
$P_\text{head}$ & $P_\text{tail}$ & $f$ & $\phi$ & $L_\text{tail}$ & $EI_\text{tail}$ \\
\hline
448 kPa & 414 kPa & 6.3 Hz & 0.14 & 53 mm & 17 mN$\cdot$m$^2$
\end{tabular}
\end{table}


The identified configuration was tested in hardware, yielding a speed slightly lower than expected. By adjusting the tail to 70mm, in line with previously observed discrepancies between computational and experimental results (as shown in Figure \ref{fig:7 Results-LengthVStiffness}(c)), we achieved a swimming speed of $21.7 \pm 0.4$, which is very close to the theoretical prediction. A comparison between other experimental observations and computational predictions for the case with optimized gait and body parameters is given in Table \ref{Table-performanceComparison}, where close agreement is obtained. 
\begin{table}
    \centering
    \caption{Performance Parameter Comparison Model vs Experimental}
\begin{tabular}{|cc||c|c|}
\hline
Value                & Symbol & \begin{tabular}[c]{@{}c@{}}LAEBT-Beam\\  Model\end{tabular} & \begin{tabular}[c]{@{}c@{}}Experimental \\ SLIDER Robot\end{tabular} \\ \hline
Free Swimming Speed  & $U$      & 23.2 cm/s                                                        & 22 cm/s                                                                   \\ \hline
Body Wavelength      & $\lambda$   & 28 cm                                                       & 26.5 cm                                                              \\ \hline
Traveling wave Speed & $V$      & 177 cm/s                                                    & 167 cm/s                                                             \\ \hline
Froude Efficiency   & $\eta_f=(U+V)/2V$ & 0.57                                                       & 0.57                                                                \\ \hline
Force Efficiency     & $\eta$    & 0.31                                                       & N/A                                                                  \\ \hline
Tail Lat. Excursion  & $A_t$         &3.6 cm                                                      &    3.9 cm \\ \hline 
Strouhal Number      & $St = (2\,A_t\,f)/U$      &1.96                                                       & 2.12\\ \hline
Tail Slope           & $A_t\,/\,(\lambda/4)$ &0.57&0.59\\ \hline
\end{tabular}
\label{Table-performanceComparison}
\end{table}

\begin{figure}
  \centering
    \includegraphics[width= 1.0\textwidth]{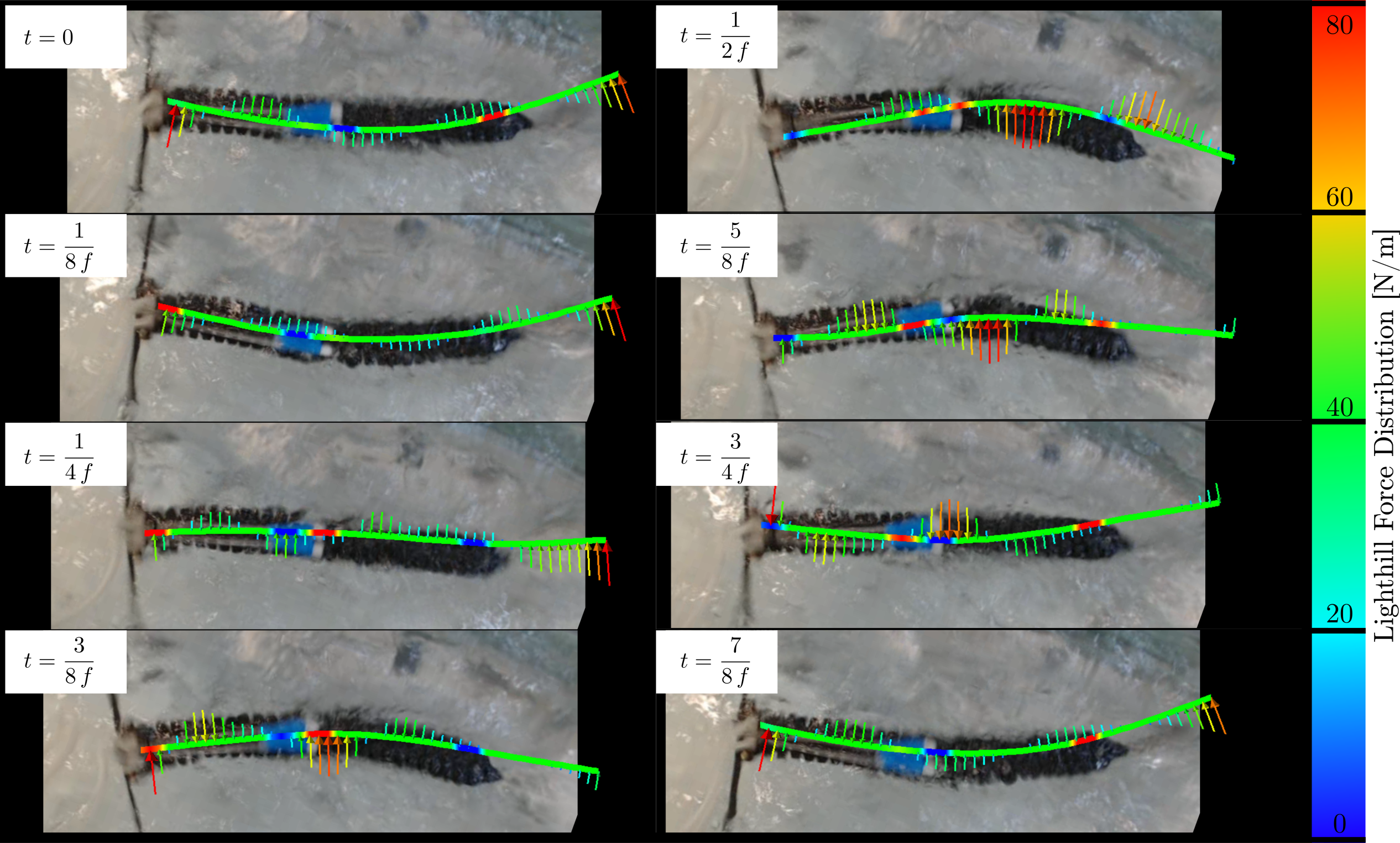}
    \caption{Experimental video frames and computational data overlay for the optimized swimmer. Arrows depict hydrodynamic force distribution. Red and blue bars along the beam length represent the moment distribution. \requestedChange{Supplementary video available.}}
    \label{fig8: Optimized Swimmer}
\end{figure}

Frames of the tracking video from the gantry for a converged experiment are shown in Figure \ref{fig8: Optimized Swimmer}. Here, the left column is the first half of the stride, and the right column is the second half of the stride. The gait is anguilliform with a $L$/$\lambda$ ratio of 1.4. The LAEBT-Beam model centerline body positions are laid over the experimental video in green. The net bending moment from the controlled internal pressure is also shown in red and blue. The hydrodynamic environmental force distribution is depicted as a vector along the body, revealing that the tail thrust is maximum at phase = 0 and at $\frac{1}{2}\,f$. 

\begin{figure}
  \centering
    \includegraphics[width= 1.0\textwidth]{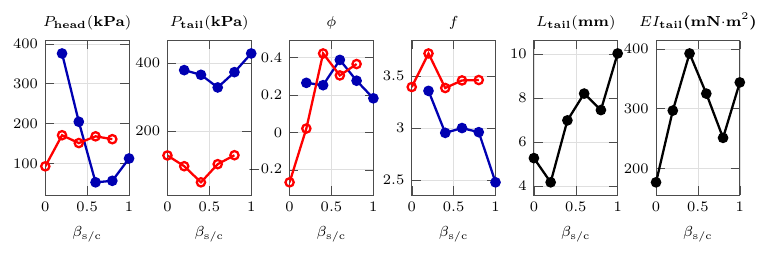}
    \caption{\requestedChangeTWO{The plot illustrates the evolution of optimal tail parameters ($L_\text{tail}$, $EI_\text{tail}$) and actuation variables (pressure $P$, frequency $f$ and phase $\phi$) as a function of the swimming-to-climbing weight $\beta_{s/c}$ ($\beta_{s/c}=1$ is for the pure swimming and $\beta_{s/c}=0$ is for the pure climbing tasks). Blue markers denote swimming-specific parameters, red markers denote climbing-specific parameters, and black lines track the geometric evolution of the tail. }}
    \label{fig9:Optmultimodal}
\end{figure}

\requestedChangeTWO{
\subsection{Multimodal Co-Optimization}
While a comprehensive study of multimodal optimization lies beyond our current scope, we demonstrate how this methodology can be applied to SLIDER topologies to maximize efficiency across combined swimming and climbing tasks. For this analysis, we set the climbing robot's rigid head link to $l_p = 3$ cm. To ensure stability and mitigate the high dynamic response associated with large body deformations during climbing, actuation pressures for the front and rear chambers ($P_\text{head}, P_\text{tail}$ ) are capped at 172.3 kPa (25 Psi).

The multimodal genetic algorithm utilizes a combined objective function:
\begin{equation}
\eta=\beta_\text{s/c}\left(\frac{\mathcal{E}^{\text{co-opt}}_\text{swim}}{\mathcal{E}^{\text{only}}_\text{swim}}\right)+(1-\beta_\text{s/c})\left(\frac{\mathcal{E}^{\text{co-opt}}_\text{clim}}{\mathcal{E}^{\text{only}}_\text{clim}}\right)
\end{equation}
where $0<\beta_\text{s/c}<1$ weights the relative importance of swimming versus climbing. The ratio $\mathcal{E}^{\text{co-opt}} / \mathcal{E}^{\text{only}}$ normalizes the co-optimized performance against the best value achieved in single-mode optimization. For swimming, $\mathcal{E}_\text{swim}$ is the ratio of useful work (${\bar{F}_T}\,\bar{U}_\text{swim}$) to mean power expenditure, averaged over 50 steady-state cycles. For climbing, we also assumed  $\mathcal{E}_\text{clim}$ is formed based on average climbing efficiency, calculated as useful work (${m\,g} \,\bar{U}_\text{clim}$) over mean power over 15 steady-state cycles. Figure \ref{fig9:Optmultimodal} illustrates how the optimal tail design and actuation parameters adapt depending on the balance between swimming and climbing efficiency. Because these optimal values shift rapidly within a specific range of $\beta_{s/c}$, a more detailed investigation into multimodal scenarios is warranted. Addressing this in future work will be critical for ensuring the robot dynamically delivers optimal performance based on the anticipated length of its climbing and swimming missions.}

\section{Discussion}
\label{Sec:04-conclusion}
Figures \ref{fig5:one Hz dynamics} and \ref{fig6:4p8 Hz Body Dynamics} illustrate the strong alignment between simulation and experimental results in the head-clamped case. The \SI{1}{\Hz} case (figure \ref{fig5:one Hz dynamics})   represents a backward swimming gait. The model captures the negative net thrust with good fidelity. The small mismatches that do appear are due to tip angle mismatch and higher-order hydrodynamics such as vortex separation.

The \SI{4.8}{\Hz} actuation case in Figure \ref{fig6:4p8 Hz Body Dynamics} demonstrates an improved match between experimental and model results, with only a 7\% error in net thrust. Despite some asymmetry, thrust minima and maxima align well. Both the \SI{1.0}{\Hz} and \SI{4.8}{\Hz} cases exhibit oscillations from the pressure manifold controller. However, the maxima and minima appear at the expected point in the stride.

The results in Figure \ref{fig4:Results-FreqSweep} show a strong match between experimental and simulation data. At low frequencies, fluid behavior is drag-dominated with negligible inertial forces, reflected in a zero added mass and a drag coefficient of 4.0. At higher frequencies, thrust increases due to greater resistive forces and dominant inertial effects, with an added mass of 1.0 and a drag coefficient of 3.0. These findings align with Kuelegan and Carpenter, who reported comparable values for drag and added mass \cite{keulegan1956forces}.

The tethered swimming results in Figure \ref{fig:7 Results-LengthVStiffness} (a) and (c) show strong agreement between the computational and experimental data. Optimal performance is observed for tail configurations similar to those identified in both methods, although the computational model's peak performance occurs with a slightly shorter tail than in the experiments. A closer match is found in the actuation phase shift results, which is particularly valuable for hardware design, as achieving a $180^\circ$ or $0.5$ phase shift is relatively simple in hardware with five-way valves but is shown here to perform poorly compared to the $0.2$ phase shift.
\requestedChange{
Using a genetic algorithm for global optimization, we improved swimming speed to approximately 22 cm/s. This is a competitive performance when compared specifically to other long, slender soft robots with multimodal capabilities, including climbing \cite{chi2022snapping,gallentine2022multimodal}. 
}

\requestedChange{
SLIDER’s optimal gait shows a notable Strouhal number and tail kinematics. Prior work found a tail slope of 1 effective for speed~\cite{anastasiadis2023identification}, though that study constrained kinematics within motor limits. In contrast, our model accounts for actuation bandwidth, allowing the optimizer to select a gait with a reduced slope of 0.59 and a frequency of \SI{6.3}{\Hz}, effectively trading tail amplitude for higher beat frequency. While efficient cruise swimming typically occurs near a Strouhal number of 0.3~\cite{taylor2003flying}, higher values, such as up to 2 in accelerating pike, are connected to high thrust generation~\cite{conte2010fast}. This implies that SLIDER’s high Strouhal number reflects optimization for speed over efficiency.
}

The genetic algorithm's optimal design increases the tail beat frequency by 20\% and selects the maximum stiffness allowed by the optimizer's boundaries. The chosen phase delay of $0.14$ aligns with the trends observed in our parameter variation study shown in Figure \ref{fig:7 Results-LengthVStiffness}. This indicates that the fundamental speed trends in Figure \ref{fig:7 Results-LengthVStiffness} remain consistent across the parameter range relevant for swimming performance. \requestedChangeTWO{ Finally, Figure \ref{fig9:Optmultimodal} reveals how balancing the optimization weights between multimodal tasks—namely climbing and swimming—results in different optimal geometries and actuation strategies.}

\section{Future Work and Conclusions}
This work is inspired by the Pacific lamprey, a long, slender fish known for its ability to swim efficiently over long distances and climb wet vertical surfaces like weirs and waterfalls. Using LAEBT and nonlinear beam theory, we developed a fast coupled model for soft robotic swimming. The model was validated through experiments, showing a strong match between computational predictions and experimental results.

In this work, we use an efficient computational model to co-optimize the body design and control for a soft robot. We calibrate our model using net thrust, time-varying thrust, and recoil forces. The model accurately captures inertia, body compliance, internal damping, and actuation pressure effects. At low frequencies, fluid forces are resistive, dominated by drag, while at higher frequencies, reactive inertial forces take over. Both the drag coefficient ($C_d$) and added mass term ($m_a$) agree with prior experiments on oscillating plates. This approach improved SLIDER's swimming performance by 178\%.\requestedChangeTWO{ In addition, we showed how to formulate the optimization task to design an optimal robot for multiple tasks required for multimodal locomotion. }

\requestedChangeTWO{A critical direction for future inquiry lies in defining robust multimodal design criteria and formulating highly detailed objective functions.} Modeling transitions from swimming to climbing is another key future direction. This would enhance our understanding of the Pacific lamprey and aid in designing robots that mimic its movement. It could help identify optimal transition gaits. Additionally, the current LAEBT theory does not capture certain fluid dynamics, such as complex vortex separation along the body before the tail tip. While this vortex formation is significant in some biological systems, such as tuna tails, its net effect is minimal for streamlined swimmers. However, SLIDER's body shape may still induce vortices along its edges, whose effects are described in a reduced-order fashion using a quadratic resistive drag term. Future work will aim to explicitly model these interactions using coupled nonlinear drag and added-mass terms. Recent studies suggest that reduced-order models can be derived from high-fidelity simulations, offering a systematic approach to establishing this fast computational model, which will be explored in upcoming research.

\section*{Acknowledgments}
We thank Anne Grote (US Fish and Wildlife Service), Kinsey Frick (NOAA Fisheries), Ralph Lampmann (Yakama Nation FRMP, Pacific Lamprey Project), and their colleagues for providing the Lamprey climbing video. This work was funded by the National Science Foundation’s Expanding Frontiers in Research and Innovation Program (grant number 1935278).

\bibliographystyle{plain}
\bibliography{Brian_Bibliography}

\begin{thebibliography}{10}

\bibitem{anastasiadis2023identification}
Alexandros Anastasiadis, Laura Paez, Kamilo Melo, Eric~D Tytell, Auke~J Ijspeert, and Karen Mulleners.
\newblock Identification of the trade-off between speed and efficiency in undulatory swimming using a bio-inspired robot.
\newblock {\em Scientific Reports}, 13(1):15032, 2023.

\bibitem{beal2006passive}
DN~Beal, FS~Hover, MS~Triantafyllou, JC~Liao, and George~V Lauder.
\newblock Passive propulsion in vortex wakes.
\newblock {\em Journal of Fluid Mechanics}, 549:385--402, 2006.

\bibitem{bongard2006resilient}
Josh Bongard, Victor Zykov, and Hod Lipson.
\newblock Resilient machines through continuous self-modeling.
\newblock {\em Science}, 314(5802):1118--1121, 2006.

\bibitem{opencv_library}
G.~Bradski.
\newblock {The OpenCV Library}.
\newblock {\em Dr. Dobb's Journal of Software Tools}, 2000.

\bibitem{candelier2011three}
Fabien Candelier, Fr{\'e}d{\'e}ric Boyer, and Alban Leroyer.
\newblock Three-dimensional extension of lighthill's large-amplitude elongated-body theory of fish locomotion.
\newblock {\em Journal of Fluid Mechanics}, 674:196--226, 2011.

\bibitem{cheng1998continuous}
J-Y Cheng, TJ~Pedley, and JD~Altringham.
\newblock A continuous dynamic beam model for swimming fish.
\newblock {\em Philosophical Transactions of the Royal Society of London. Series B: Biological Sciences}, 353(1371):981--997, 1998.

\bibitem{chi2022snapping}
Yinding Chi, Yaoye Hong, Yao Zhao, Yanbin Li, and Jie Yin.
\newblock Snapping for high-speed and high-efficient butterfly stroke--like soft swimmer.
\newblock {\em Science Advances}, 8(46):eadd3788, 2022.

\bibitem{conte2010fast}
J~Conte, Yahya Modarres-Sadeghi, MN~Watts, Franz~Stephen Hover, and Michael~S Triantafyllou.
\newblock A fast-starting mechanical fish that accelerates at 40 ms- 2.
\newblock {\em Bioinspiration \& biomimetics}, 5(3):035004, 2010.

\bibitem{cruz2020soft}
Christyan Cruz~Ulloa, Silvia Terrile, and Antonio Barrientos.
\newblock Soft underwater robot actuated by shape-memory alloys “jellyrobcib” for path tracking through fuzzy visual control.
\newblock {\em Applied Sciences}, 10(20):7160, 2020.

\bibitem{eloy2012origin}
Christophe Eloy, Nicolas Kofman, and Lionel Schouveiler.
\newblock The origin of hysteresis in the flag instability.
\newblock {\em Journal of fluid mechanics}, 691:583--593, 2012.

\bibitem{feng2020body}
Hui Feng, Yi~Sun, Peter~A Todd, and Heow~Pueh Lee.
\newblock Body wave generation for anguilliform locomotion using a fiber-reinforced soft fluidic elastomer actuator array toward the development of the eel-inspired underwater soft robot.
\newblock {\em Soft Robotics}, 7(2):233--250, 2020.

\bibitem{gallentine2022multimodal}
James Gallentine, Eric~J. Barth, Kevin Galloway, Brian Van~Stratum, Jonathan Clark, and Kourosh Shoele.
\newblock A multimodal climbing-swimming soft robotic lamprey.
\newblock In {\em Proceedings of the 2022 Bath/ASME Symposium on Fluid Power and Motion Control, FPMC2022}, 2022.

\bibitem{katoch2021review}
Sourabh Katoch, Sumit~Singh Chauhan, and Vijay Kumar.
\newblock A review on genetic algorithm: past, present, and future.
\newblock {\em Multimedia tools and applications}, 80:8091--8126, 2021.

\bibitem{keulegan1956forces}
Garbis~H Keulegan, Lloyd~H Carpenter, et~al.
\newblock {\em Forces on cylinders and plates in an oscillating fluid}.
\newblock National Bureau of Standards, 1956.

\bibitem{leclercq2018vortex}
Tristan Leclercq and Emmanuel de~Langre.
\newblock Vortex-induced vibrations of cylinders bent by the flow.
\newblock {\em Journal of Fluids and Structures}, 80:77--93, 2018.

\bibitem{lighthill1971large}
Michael~James Lighthill.
\newblock Large-amplitude elongated-body theory of fish locomotion.
\newblock {\em Proceedings of the Royal Society of London. Series B. Biological Sciences}, 179(1055):125--138, 1971.

\bibitem{lighthill1960note}
MJ~Lighthill.
\newblock Note on the swimming of slender fish.
\newblock {\em Journal of fluid Mechanics}, 9(2):305--317, 1960.

\bibitem{liu2022fishgym}
Wenji Liu, Kai Bai, Xuming He, Shuran Song, Changxi Zheng, and Xiaopei Liu.
\newblock Fishgym: A high-performance physics-based simulation framework for underwater robot learning.
\newblock In {\em 2022 International Conference on Robotics and Automation (ICRA)}, pages 6268--6275. IEEE, 2022.

\bibitem{ma2021diffaqua}
Pingchuan Ma, Tao Du, John~Z Zhang, Kui Wu, Andrew Spielberg, Robert~K Katzschmann, and Wojciech Matusik.
\newblock Diffaqua: A differentiable computational design pipeline for soft underwater swimmers with shape interpolation.
\newblock {\em ACM Transactions on Graphics (TOG)}, 40(4):1--14, 2021.

\bibitem{mosadegh2014pneumatic}
Bobak Mosadegh, Panagiotis Polygerinos, Christoph Keplinger, Sophia Wennstedt, Robert~F Shepherd, Unmukt Gupta, Jongmin Shim, Katia Bertoldi, Conor~J Walsh, and George~M Whitesides.
\newblock Pneumatic networks for soft robotics that actuate rapidly.
\newblock {\em Advanced functional materials}, 24(15):2163--2170, 2014.

\bibitem{nava2022fast}
Elvis Nava, John~Z Zhang, Mike~Yan Michelis, Tao Du, Pingchuan Ma, Benjamin~F Grewe, Wojciech Matusik, and Robert~Kevin Katzschmann.
\newblock Fast aquatic swimmer optimization with differentiable projective dynamics and neural network hydrodynamic models.
\newblock In {\em International Conference on Machine Learning}, pages 16413--16427. PMLR, 2022.

\bibitem{nguyen2021evaluation}
Dinh~Quang Nguyen et~al.
\newblock Evaluation on swimming efficiency of an eel-inspired soft robot with partially damaged body.
\newblock In {\em 2021 IEEE 4th International Conference on Soft Robotics (RoboSoft)}, pages 289--294. IEEE, 2021.

\bibitem{nguyen2020kinematic}
Dinh~Quang Nguyen and Van~Anh Ho.
\newblock Kinematic evaluation of a series of soft actuators in designing an eel-inspired robot.
\newblock In {\em 2020 IEEE/SICE International Symposium on System Integration (SII)}, pages 1288--1293, 2020.

\bibitem{nguyen2021anguilliform}
Dinh~Quang Nguyen and Van~Anh Ho.
\newblock Anguilliform swimming performance of an eel-inspired soft robot.
\newblock {\em Soft Robotics}, 2021.

\bibitem{peskin2002immersed}
Charles~S Peskin.
\newblock The immersed boundary method.
\newblock {\em Acta numerica}, 11:479--517, 2002.

\bibitem{porez2014improved}
Mathieu Porez, Fr{\'e}d{\'e}ric Boyer, and Auke~Jan Ijspeert.
\newblock Improved lighthill fish swimming model for bio-inspired robots: Modeling, computational aspects and experimental comparisons.
\newblock {\em The International Journal of Robotics Research}, 33(10):1322--1341, 2014.

\bibitem{sarpkaya1976vortex}
Turgut Sarpkaya.
\newblock Vortex shedding and resistance in harmonic flow about smooth and rough circular cylinders.
\newblock {\em BOSS-76}, 1:220--235, 1976.

\bibitem{sarpkaya2010wave}
Turgut Sarpkaya.
\newblock {\em Wave forces on offshore structures}.
\newblock Cambridge university press, 2010.

\bibitem{shoele2016flutter}
Kourosh Shoele and Rajat Mittal.
\newblock Flutter instability of a thin flexible plate in a channel.
\newblock {\em Journal of Fluid Mechanics}, 786:29--46, 2016.

\bibitem{taylor1952analysis}
Geoffrey~Ingram Taylor.
\newblock Analysis of the swimming of long and narrow animals.
\newblock {\em Proceedings of the Royal Society of London. Series A. Mathematical and Physical Sciences}, 214(1117):158--183, 1952.

\bibitem{taylor2003flying}
Graham~K Taylor, Robert~L Nudds, and Adrian~LR Thomas.
\newblock Flying and swimming animals cruise at a strouhal number tuned for high power efficiency.
\newblock {\em Nature}, 425(6959):707--711, 2003.

\bibitem{tjavaras1996dynamics}
Athanassios~Andreas Tjavaras.
\newblock {\em The dynamics of highly extensible cables}.
\newblock PhD thesis, Massachusetts Institute of Technology, 1996.

\bibitem{VanStratum2023Pressure}
Brian Van~Stratum.
\newblock {Soft Snake Pressure Control}.
\newblock \url{https://github.com/bvanstratum/SoftSnakePressureControl}, October 2023.

\bibitem{vanstratum2023Effects}
Brian Van~Stratum, Jonathan Clark, and Kourosh Shoele.
\newblock Effect of internal damping on locomotion in frictional environments.
\newblock {\em Phys. Rev. E}, 107:054406, May 2023.

\bibitem{wendroff1959centered}
Burton Wendroff.
\newblock On centered difference equations for hyperbolic systems.
\newblock {\em Journal of the Society for Industrial and Applied Mathematics}, 8(3):549--555, 1960.

\bibitem{zhu2011numerical}
Q~Zhu, M~Moser, and P~Kemp.
\newblock Numerical analysis of a unique mode of locomotion: vertical climbing by pacific lamprey.
\newblock {\em Bioinspiration \& Biomimetics}, 6(1):016005, 2011.

\bibitem{zhu2008propulsion}
Qiang Zhu and Kourosh Shoele.
\newblock Propulsion performance of a skeleton-strengthened fin.
\newblock {\em Journal of Experimental Biology}, 211(13):2087--2100, 2008.

\end{thebibliography}
\end{document}